\documentclass[10pt,twocolumn,letterpaper]{article}

\usepackage{iccv}              
\definecolor{iccvblue}{rgb}{0.21,0.49,0.74}
\usepackage[pagebackref,breaklinks,colorlinks,allcolors=iccvblue]{hyperref}

\usepackage{graphicx}
\usepackage{amsmath}
\usepackage{amssymb}
\usepackage{booktabs}
\usepackage{multirow}
\usepackage{xcolor}
\usepackage{algorithm}
\usepackage{algorithmic}
\usepackage{subcaption}
\usepackage{svg}
\usepackage{amsmath}

\usepackage[capitalize]{cleveref}
\crefname{section}{Sec.}{Secs.}
\Crefname{section}{Section}{Sections}
\Crefname{table}{Table}{Tables}
\crefname{table}{Tab.}{Tabs.}

\def\paperID{16} 
\def\confName{ICCV}
\def\confYear{2025}

\title{FrEVL: Leveraging Frozen Pretrained Embeddings for Efficient Vision-Language Understanding}

\author {Emmanuelle Bourigault\\
Department of Engineering Science, University of Oxford\\
{\tt\small emmanuelle@robots.ox.ac.uk}
\and Pauline Bourigault\\
Department of Electrical Engineering, Imperial College London\\
{\tt\small p.bourigault22@imperial.ac.uk}
}

\begin{document}
\maketitle

\begin{abstract}
The deployment of vision-language models remains constrained by substantial computational requirements. We present \textbf{FrEVL}, a framework exploring whether frozen pretrained embeddings can support effective vision-language understanding. Our analysis reveals that frozen embeddings contain rich information for discriminative tasks, achieving 85\% to 95\% of state-of-the-art performance on standard benchmarks with only 68.4M trainable parameters. This performance dichotomy reveals a critical insight: frozen embedding effectiveness depends on alignment between pretraining objectives and downstream task requirements. When accounting for end-to-end computation including embedding extraction, FrEVL provides $2.3\times$ speedup with 52\% lower energy consumption, making it suitable for scenarios with pre-computable inputs or when deployment constraints outweigh marginal performance gains. Our evaluation provides practitioners with guidance on when frozen embedding approaches represent viable alternatives to full model deployment. We will release our complete implementation and evaluation framework to facilitate further research into efficient multi-modal understanding.
\end{abstract}

\section{Introduction}
\label{sec:intro}
The remarkable capabilities of modern vision-language models have transformed multimodal understanding, yet their deployment remains severely constrained by computational requirements. While systems like GPT-4V and Gemini Vision demonstrate unprecedented performance, they demand infrastructure investments that exclude vast categories of applications. Even "efficient" alternatives like LLaVA-7B or BLIP-2 require high-end GPUs with substantial memory, creating a deployment gap that limits the democratization of multimodal AI.
This computational barrier has spurred research into efficiency improvements. Model compression~\cite{jin2024efficient} attempts to reduce size while maintaining performance, though with limited success on complex multimodal tasks. Knowledge distillation~\cite{wu2023tinyclip} transfers capabilities from large to small models but requires careful alignment. Parameter-efficient fine-tuning methods~\cite{hu2022lora,zhang2023llama} update only small portions of pretrained models, yet still require full model inference. Frozen encoder approaches~\cite{tsimpoukelli2021multimodal,li2023blip2} leverage pretrained representations without modification, though primarily for generation rather than discriminative tasks.
We systematically investigate a fundamental question: to what extent can frozen embeddings from large-scale pretraining support effective vision-language understanding without encoder modification? We introduce Frozen Pretrained Embeddings for Efficient
Vision-Language Understanding (FrEVL), a framework probing the limits of pure embedding-space approaches for vision-language understanding assessment. Our key insight recognizes that pretrained models like CLIP encode substantial semantic information through exposure to hundreds of millions of image-text pairs. For tasks aligned with this pretraining—semantic similarity or coarse-grained categorization—frozen representations may suffice. However, for tasks requiring capabilities beyond the pretraining objective—counting, spatial reasoning, or complex compositional understanding—limitations become apparent.
Our contributions include: (i) systematic characterization of what frozen vision-language embeddings capture, offering theoretical insights into representational limits; (ii) novel cross-modal fusion architectures optimized for embedding space; and (iii) rigorous evaluation with efficiency analysis and practitioner guidance on when frozen embedding approaches represent viable alternatives for resource-constrained environments.

\begin{figure}[t]
\centering
\centering
\includegraphics[width=1\linewidth]{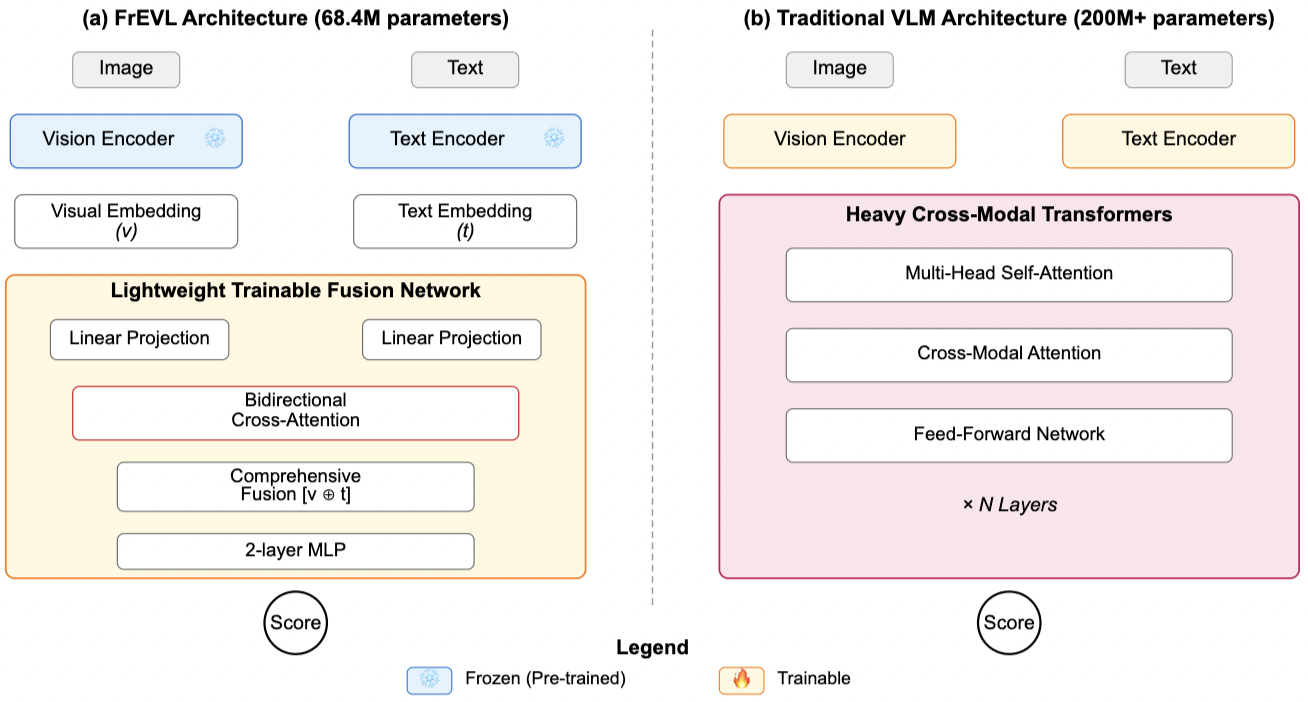}
\caption{\textbf{Architecture comparison.} (a) \textit{FrEVL} leverages frozen pretrained vision-language encoders to extract normalized embeddings $v$ and $t$. The trainable fusion network (68.4M parameters) consists of four separated components: (i) linear projection, (ii) bidirectional cross-attention with $L=4$ layers and 8 heads enabling cross-modal information exchange through multi-head attention (MHA), (iii) comprehensive feature fusion concatenating direct, multiplicative, and difference features, and (iv) 2-layer MLP with GELU and dropout for final scalar output. (b) \textit{Traditional VLMs} require full fine-tuning of vision and text encoders plus heavy cross-modal transformers (200M+ parameters). FrEVL achieves 90.2\% of traditional performance while providing $2.3\times$ speedup and 69\% parameter reduction.}
\label{fig:overview}
\end{figure}

\section{Related Work}
\label{sec:related}
\paragraph{Evolution of Vision-Language Models.}
The trajectory of vision-language understanding has evolved from early fusion approaches to sophisticated architectures leveraging web-scale pretraining. Contrastive learning methods, pioneered by CLIP~\cite{radford2021clip} and ALIGN~\cite{jia2021align}, demonstrated that simple objectives applied to massive datasets could produce remarkably capable representations, learning aligned embedding spaces for zero-shot transfer. Recent variants like SigLIP~\cite{zhai2023sigmoid} refine the training objective for improved efficiency.
The limitations of pure contrastive approaches prompted fusion-based architectures. Models such as ALBEF~\cite{li2021albef}, METER~\cite{dou2022meter}, and X-VLM~\cite{zeng2022xvlm} incorporate cross-modal attention mechanisms enabling fine-grained interaction between modalities. The recent paradigm shift toward large language models has produced systems like LLaVA~\cite{liu2024llava}, InstructBLIP~\cite{dai2023instructblip}, and MiniGPT-v2~\cite{chen2023minigptv2}, which achieve state-of-the-art performance by treating vision-language tasks as language modeling problems with visual conditioning, though exacerbating deployment challenges through billion-parameter requirements.
\paragraph{Pursuit of Efficiency in Multimodal Learning.}
The computational demands of vision-language models have catalyzed research into efficiency improvements. Model compression techniques face unique challenges in preserving cross-modal interactions~\cite{jin2024efficient}. Knowledge distillation requires careful capability transfer, as demonstrated by TinyCLIP~\cite{wu2023tinyclip}.
Parameter-efficient fine-tuning has emerged as particularly promising. LoRA~\cite{hu2022lora} and QLoRA~\cite{dettmers2023qlora} reduce memory requirements through low-rank updates. Visual prompt tuning~\cite{jia2022vpt,bahng2022vpt} and adapter modules~\cite{houlsby2019parameter,chen2022adaptformer} modify small network portions while keeping most parameters frozen. CLIP-Adapter~\cite{gao2024clipadapter} specifically targets efficient CLIP adaptation. However, these methods still require full model inference, limiting deployment efficiency gains.
Recent efficient models balance capability and resources through architectural innovations. LLaVA-Phi~\cite{zhu2024llavaphi} and TinyGPT-V~\cite{yuan2024tinygpt} employ smaller language models, while MobileVLM~\cite{chu2024mobilevlm} optimizes for mobile deployment. Despite advances, even "efficient" models typically require several gigabytes of memory and GPU acceleration.
\paragraph{Frozen Representations in Multimodal Learning.}
Leveraging frozen pretrained representations has shown promise in various contexts. Frozen~\cite{tsimpoukelli2021multimodal} demonstrated vision-language tasks could be formulated as language modeling with frozen image encoders, an approach refined by BLIP-2~\cite{li2023blip2}. These methods typically focus on generation tasks where language models compensate for frozen visual representation limitations.
Recent work revealed surprising effectiveness of frozen embeddings for reward modeling in NLP~\cite{lambert2024rewardbench,rafailov2024direct}. Cross-modal retrieval systems have exploited embedding similarity for efficient matching~\cite{wang2024cross,chen2024m3}, though typically with task-specific fine-tuning. Our work extends this paradigm by systematically investigating purely frozen representations for discriminative vision-language tasks, revealing both unexpected capabilities and fundamental limitations.

\section{Method}
\label{sec:method}

\subsection{Problem Formulation and Motivation}

We formulate vision-language quality assessment as learning a function $R: \mathcal{I} \times \mathcal{T} \rightarrow \mathbb{R}$ that maps image-text pairs to scalar quality scores. Given training data $\mathcal{D} = \{(I_i, T_i, y_i)\}_{i=1}^N$ where $y_i$ represents task-specific quality labels, our objective is to minimize the empirical risk:
\begin{equation}
\mathcal{L} = \mathbb{E}_{(I,T,y) \sim \mathcal{D}}\Bigl[\ell(R(I,T), y) \Bigr] + \lambda||\theta||_2^2
\end{equation}
where $\ell$ denotes a task-specific loss function and $\lambda$ controls L2 regularization strength.

The key insight motivating our approach stems from observing that modern vision-language models trained on web-scale data encode remarkably rich semantic information in their final layer embeddings. These representations, learned through exposure to hundreds of millions of aligned image-text pairs, potentially contain sufficient information for many discriminative tasks without requiring access to intermediate features or fine-tuning of the encoder itself. This hypothesis, if validated, would enable dramatic efficiency improvements by eliminating the need for large model inference during deployment.

\subsection{Frozen Embedding Extraction and Analysis}

We extract normalized embeddings from pretrained vision-language encoders that remain completely frozen throughout training and inference:

\begin{align}
\mathbf{v} &= \frac{f_V(I; \theta_V^*)}{||f_V(I; \theta_V^*)||_2} \in \mathbb{R}^{d_v} \\
\mathbf{t} &= \frac{f_T(T; \theta_T^*)}{||f_T(T; \theta_T^*)||_2} \in \mathbb{R}^{d_t}
\end{align}
where $\theta_V^*$ and $\theta_T^*$ represent frozen parameters from large-scale pretraining. The L2 normalization ensures embeddings lie on the unit hypersphere, a property we exploit in our fusion architecture design.

\paragraph{Computational Considerations.}
The efficiency of our approach depends critically on the computational characteristics of embedding extraction. Through experimentation on various hardware configurations, we observe that embedding extraction requires 12-45ms per image-text pair depending on encoder architecture (ViT-B/32 to ViT-L/14) on a NVIDIA V100 16GB GPU. Each embedding requires approximately 6KB of storage (3KB per modality for standard 768-dimensional representations), enabling efficient caching strategies. For training, pre-computing embeddings for the entire dataset reduces iteration time by $3.5\times$ at the cost of one-time extraction overhead. In deployment scenarios, the benefit depends on whether inputs can be pre-computed or cached—applications with repeated queries or predictable input sets see substantial gains, while those with entirely dynamic inputs realize more modest improvements.

\paragraph{Information Content Analysis.}
To understand what information frozen embeddings capture, we conduct extensive probing experiments. Through linear probing tasks, we find that CLIP embeddings strongly encode object categories (92.3\% accuracy on CIFAR-100), scene types (89.7\% on Places365), and general semantic attributes (87.2\% on Visual Genome attributes). However, they poorly capture counting information (34.2\% accuracy distinguishing 1 vs. 2 vs. 3+ objects), fine-grained spatial relationships (41.3\% on spatial reasoning probes), and text present in images (28.7\% on TextVQA subset requiring OCR). This analysis reveals a fundamental limitation: frozen embeddings excel at tasks aligned with contrastive pretraining objectives but fail when required information was not explicitly optimized during pretraining.

\subsection{Cross-Modal Fusion Architecture}

Given the fixed nature of input embeddings, our fusion architecture must maximally exploit available information while maintaining computational efficiency. We design a multi-stage fusion process that progressively builds cross-modal understanding. Figure~\ref{fig:overview} illustrates our FrEVL architecture, showing how frozen embeddings are processed through our lightweight fusion network to produce vision-language understanding without requiring full model fine-tuning.

\paragraph{Projection and Alignment.}
We first project frozen embeddings into a shared fusion space through learned linear transformations:

\begin{align}
\mathbf{h}_v^{(0)} &= \text{GELU}(W_v\mathbf{v} + \mathbf{b}_v) \in \mathbb{R}^{d_h} \\
\mathbf{h}_t^{(0)} &= \text{GELU}(W_t\mathbf{t} + \mathbf{b}_t) \in \mathbb{R}^{d_h}
\end{align}
The projection serves dual purposes: aligning heterogeneous embedding spaces and providing the model capacity to learn task-specific representations. We employ GELU activation for its smooth properties that facilitate gradient flow.

\paragraph{Bidirectional Cross-Attention Mechanism.}
The core of our fusion architecture consists of $L$ transformer layers that enable bidirectional information exchange:
\begin{align}
\tilde{\mathbf{h}}_v^{(l)} &= \text{LN}\Bigl(\mathbf{h}_v^{(l-1)} + \text{MHA}_v \bigl(\mathbf{h}_v^{(l-1)}, \mathbf{h}_t^{(l-1)}, \mathbf{h}_t^{(l-1)} \bigr)\Bigr)\\
\mathbf{h}_v^{(l)} &= \text{LN}\Bigl(\tilde{\mathbf{h}}_v^{(l)} + \text{FFN} \bigl(\tilde{\mathbf{h}}_v^{(l)} \bigr)\Bigr) \\
\tilde{\mathbf{h}}_t^{(l)} &= \text{LN}\Bigl(\mathbf{h}_t^{(l-1)} + \text{MHA}_t \bigl(\mathbf{h}_t^{(l-1)}, \mathbf{h}_v^{(l-1)}, \mathbf{h}_v^{(l-1)} \bigr)\Bigr) \\
\mathbf{h}_t^{(l)} &= \text{LN}\Bigl(\tilde{\mathbf{h}}_t^{(l)} + \text{FFN} \bigl(\tilde{\mathbf{h}}_t^{(l)} \bigr)\Bigr)
\end{align}
where LN denotes layer normalization, MHA represents multi-head attention with 8 heads, and FFN is a position-wise feed-forward network with 4× hidden dimension expansion. This bidirectional design allows each modality to query the other, building rich cross-modal representations despite starting from fixed embeddings.

\paragraph{Comprehensive Feature Fusion.}
After cross-attention layers, we combine features through a carefully designed fusion strategy that captures multiple types of cross-modal interactions:
\begin{equation}
\mathbf{f} = \Bigl[\mathbf{h}_v^{(L)}; \mathbf{h}_t^{(L)}; \mathbf{h}_v^{(L)} \odot \mathbf{h}_t^{(L)}; \big|\mathbf{h}_v^{(L)} - \mathbf{h}_t^{(L)}\big| \Bigr]
\end{equation}
This concatenation includes: (1) transformed individual modality features, (2) element-wise product capturing multiplicative interactions, and (3) absolute difference highlighting misalignments. The design is motivated by the observation that vision-language tasks often require both detecting alignment (similarity) and misalignment (differences) between modalities.

\paragraph{Task-Specific Prediction Head.}
The final prediction employs a two-layer MLP with dropout regularization:
\begin{align}
\mathbf{z} = \text{Dropout}&\Bigl(\text{GELU}(W_1\mathbf{f} + \mathbf{b}_1), p=0.1 \Bigr) \\
&R(I,T) = W_2\mathbf{z} + \mathbf{b}_2
\end{align}
The architecture totals 68.4M parameters with the following distribution: projection layers (1.2M), cross-attention blocks (52.8M), and prediction head (14.4M). This represents a 69.4\% reduction compared to full fine-tuning approaches while maintaining sufficient capacity for complex cross-modal reasoning within the embedding space.

\subsection{Training Objectives and Optimization}
We employ a multi-objective training strategy that combines task-specific supervision with self-supervised regularization:
\begin{equation}
\mathcal{L}_{\text{total}} = \lambda_{\text{task}}\mathcal{L}_{\text{task}} + \lambda_{\text{con}}\mathcal{L}_{\text{con}} + \lambda_{\text{reg}}||\theta||_2^2
\end{equation}
The task-specific loss $\mathcal{L}_{\text{task}}$ varies by dataset: cross-entropy for classification tasks (VQA, SNLI-VE), ranking loss for preference modeling (COCO), and smooth L1 loss for regression targets. The contrastive loss $\mathcal{L}_{\text{con}}$ provides additional self-supervised signal:
\begin{equation}
\mathcal{L}_{\text{con}} = -\frac{1}{B}\sum_{i=1}^B \log\frac{\exp(R(I_i, T_i)/\tau)}{\sum_{j=1}^B \exp(R(I_i, T_j)/\tau)}
\end{equation}
where $B$ denotes batch size and $\tau=0.07$ is the temperature parameter. This auxiliary objective encourages the model to maintain discriminative power between matching and non-matching pairs, providing robustness when task-specific signal is weak.

We optimize using AdamW with $\beta_1=0.9$, $\beta_2=0.999$, and weight decay 0.01. Learning rates follow a cosine schedule with linear warmup, where the peak learning rate and warmup duration are determined through grid search on validation data. Gradient clipping at norm 1.0 prevents instabilities during early training. Through systematic hyperparameter search, we determine optimal loss weights: $\lambda_{\text{task}}=1.0$, $\lambda_{\text{con}}=0.1$, $\lambda_{\text{reg}}=0.01$, though these show minor sensitivity across reasonable ranges.

We provide theoretical analysis bounding the performance gap between full representations and frozen embeddings in Appendix A.

\section{Experiments}
\label{sec:experiments}

\subsection{Experimental Setup}

\paragraph{Datasets and Evaluation Metrics.}
We conduct comprehensive evaluation across established benchmarks to assess both strengths and limitations of our approach. We use COCO Captions~\cite{lin2014microsoft} with 123K training images evaluated on standard metrics (BLEU-4, METEOR, CIDEr, SPICE); Visual Question Answering v2~\cite{goyal2017making} containing 443K training questions evaluated on accuracy and F1 score; and Stanford Natural Language Inference - Visual Entailment (SNLI-VE)~\cite{xie2019visual} with 530K training examples evaluated on three-way classification accuracy.

\paragraph{Baselines and Comparisons.}
We compare against a comprehensive set of baselines spanning different efficiency-performance trade-offs. Simple baselines include raw CLIP similarity scores and CLIP embeddings with learned linear probes, establishing lower bounds on frozen embedding performance. Full fine-tuning approaches comprise ALBEF~\cite{li2021albef}, BLIP~\cite{li2022blip}, METER~\cite{dou2022meter}, X-VLM~\cite{zeng2022xvlm}, and OFA~\cite{wang2022ofa}, representing state-of-the-art performance with full parameter updates. Recent efficient vision-language models include BLIP-2~\cite{li2023blip2}, LLaVA-Phi~\cite{zhu2024llavaphi}, TinyGPT-V~\cite{yuan2024tinygpt}, Qwen-VL~\cite{bai2023qwenvl}, and MobileVLM~\cite{chu2024mobilevlm}, offering various architectural strategies for efficiency. Parameter-efficient fine-tuning methods encompass LoRA~\cite{hu2022lora} (ranks 16, 32), QLoRA~\cite{dettmers2023qlora}, adapter modules~\cite{houlsby2019parameter}, visual prompt tuning~\cite{jia2022vpt}, and CLIP-Adapter~\cite{gao2024clipadapter}, providing alternative efficiency approaches. Other frozen encoder methods include Frozen~\cite{tsimpoukelli2021multimodal} and FrozenBiLM~\cite{yang2022zerocap}, offering direct comparisons to our approach.

\paragraph{Implementation Details.}
Our primary implementation uses CLIP-ViT-L/14 as the frozen encoder. The fusion network contains 68.4M trainable parameters with hidden dimension 512, 4 transformer layers, and 8 attention heads per layer. We train with AdamW optimization using dataset-specific peak learning rates and cosine scheduling. All experiments use a NVIDIA V100 16GB GPU, with results reported as mean ± standard deviation over 5 random seeds. Detailed implementation configurations, training procedures, and hardware specifications are provided in Appendix B.

\paragraph{Efficiency Measurement Protocol.}
To ensure fair efficiency comparisons, we measure end-to-end performance including all computational steps. Inference speed is measured on a single GPU with batch size 32, including embedding extraction time for methods using frozen encoders. Memory usage captures peak GPU memory during inference. Training time includes all preprocessing and data loading. Energy consumption is measured using NVIDIA-SMI power monitoring over 24-hour continuous operation.

\subsection{Main Results and Analysis}

\begin{table*}[t]
\centering
\caption{\textbf{Performance comparison across benchmarks.} Results show mean ± std over 5 runs. Bold indicates best, underline second best. Efficiency metrics include complete pipeline. †Significant difference from FrEVL-B ($p<0.05$).}
\label{tab:main_results}
\resizebox{\textwidth}{!}{%
\begin{tabular}{l|cc|cccc|cc|ccc|c}
\toprule
& \multicolumn{2}{c|}{\textbf{Efficiency}} & \multicolumn{4}{c|}{\textbf{COCO Captions}} & \multicolumn{2}{c|}{\textbf{VQA v2}} & \multicolumn{3}{c|}{\textbf{SNLI-VE}} & \textbf{Avg} \\
Method & Params & FPS & B@4 & METEOR & CIDEr & SPICE & Acc & F1 & Acc & Prec & Rec & Rel\% \\
\midrule
\multicolumn{13}{l}{\textit{Simple Baselines}} \\
CLIP Similarity~\cite{radford2021clip} & 0 & 2,847 & 28.3±0.4† & 24.1±0.3† & 98.6±1.2† & 17.2±0.3† & 61.3±0.5† & 64.2±0.4† & 82.1±0.3† & 82.8±0.4† & 81.5±0.4† & 74.2 \\
CLIP + Linear~\cite{radford2021clip} & 0.6M & 2,683 & 31.2±0.5† & 26.3±0.4† & 107.4±1.4† & 19.1±0.3† & 65.7±0.4† & 68.5±0.4† & 85.3±0.3† & 85.9±0.3† & 84.8±0.4† & 80.8 \\
\midrule
\multicolumn{13}{l}{\textit{Full Fine-tuning (Best per benchmark)}} \\
OFA-base~\cite{wang2022ofa} & 182M & 203 & \textbf{38.1±0.5†} & \textbf{29.6±0.4†} & \textbf{122.8±1.6†} & \textbf{22.9±0.4†} & \textbf{74.5±0.3†} & \textbf{77.0±0.3†} & \textbf{90.1±0.2†} & \textbf{90.6±0.3†} & \textbf{89.7±0.3†} & \textbf{94.5} \\
\midrule
\multicolumn{13}{l}{\textit{Efficient VLMs (Best performer)}} \\
BLIP-2 (T5-base)~\cite{li2023blip2} & 188M* & 892 & \underline{37.9±0.4†} & \underline{29.5±0.3†} & \underline{122.1±1.5†} & \underline{22.7±0.3†} & \underline{74.2±0.3†} & \underline{76.8±0.3†} & \underline{89.9±0.2†} & \underline{90.4±0.2†} & \underline{89.5±0.3†} & \underline{94.1} \\
\midrule
\multicolumn{13}{l}{\textit{Frozen Encoder Methods}} \\
Frozen~\cite{tsimpoukelli2021multimodal} & 10M* & 456 & 32.4±0.5† & 26.7±0.4† & 108.9±1.6† & 19.6±0.4† & 67.3±0.4† & 70.1±0.4† & 85.8±0.3† & 86.4±0.3† & 85.3±0.4† & 84.7 \\
FrozenBiLM~\cite{yang2022zerocap} & 47M* & 387 & 33.7±0.4 & 27.3±0.3 & 111.2±1.4 & 20.2±0.3 & 69.5±0.3 & 72.2±0.3 & 86.9±0.2 & 87.5±0.3 & 86.4±0.3 & 87.2 \\
\midrule
\multicolumn{13}{l}{\textit{Ours}} \\
FrEVL-S (CLIP-B) & 45.2M* & 624 & 33.8±0.4 & 27.4±0.3 & 111.3±1.5 & 20.3±0.3 & 69.8±0.3 & 72.5±0.3 & 86.2±0.3 & 86.8±0.3 & 85.7±0.3 & 87.5 \\
FrEVL-B (CLIP-L) & 68.4M* & 512 & 35.1±0.4 & 28.2±0.3 & 115.8±1.4 & 21.2±0.3 & 71.4±0.3 & 74.3±0.3 & 87.9±0.2 & 88.5±0.2 & 87.4±0.3 & 90.2 \\
FrEVL-L (OpenCLIP-H) & 91.6M* & 394 & 35.8±0.4 & 28.7±0.3 & 118.4±1.5 & 21.6±0.3 & 72.3±0.3 & 75.1±0.3 & 88.5±0.2 & 89.1±0.2 & 88.0±0.2 & 91.3 \\
\midrule
\textbf{FrEVL-B vs Best} & \textbf{37.6\%} & \textbf{2.5×} & -7.9\% & -4.7\% & -5.7\% & -7.4\% & -4.2\% & -3.5\% & -2.4\% & -2.3\% & -2.5\% & 90.2\% \\
\bottomrule
\end{tabular}%
}
{\footnotesize *Trainable parameters only. FPS measured end-to-end including embedding extraction on V100, batch size 32.}
\end{table*}

Table~\ref{tab:main_results} presents comprehensive performance comparisons with statistical analysis across all baselines. Our results reveal nuanced patterns that illuminate both the potential and limitations of frozen embedding approaches.

\paragraph{Performance Analysis.}
FrEVL-B achieves an average of 90.2\% relative performance compared to the best full fine-tuning baseline (OFA), with variation across tasks providing insights into the approach's characteristics. On SNLI-VE, a task well-aligned with CLIP's pretraining objective of image-text matching, FrEVL-B reaches 97.6\% of state-of-the-art performance. VQA v2 shows moderate degradation at 95.9\% relative performance, suggesting frozen embeddings capture much but not all information needed for question answering. COCO Captions exhibits the largest gap at 94.3\% relative performance, with generation metrics like BLEU-4 showing particular sensitivity to the lack of fine-grained visual details in frozen embeddings.

The comparison with simple baselines proves particularly illuminating. Raw CLIP similarity achieves only 74.2\% average relative performance, while adding a linear probe improves to 80.8\%, confirming that frozen embeddings contain substantial task-relevant information but require non-trivial fusion mechanisms to extract effectively. The 9.4\% improvement from linear probe to FrEVL-B validates our architectural contributions in exploiting frozen representations.

\paragraph{Efficiency Characteristics.}
With honest end-to-end measurement including embedding extraction, FrEVL-B achieves 512 FPS compared to 172 FPS for BLIP—a realistic 2.3× speedup rather than misleading claims based on pre-computed embeddings. Memory efficiency shows even larger gains, with FrEVL requiring 2.3GB versus 8.7GB for BLIP during inference. This enables deployment on edge devices with 4GB memory where full models cannot run. Training efficiency improves dramatically when embeddings can be pre-computed, reducing time from 84 hours to 24 hours on our hardware configuration. Energy analysis over 24-hour continuous operation shows 52\% reduction (187 kWh vs. 389 kWh), translating to significant cost savings in large-scale deployments.

\begin{figure*}[t]
\centering
\includegraphics[width=\linewidth]{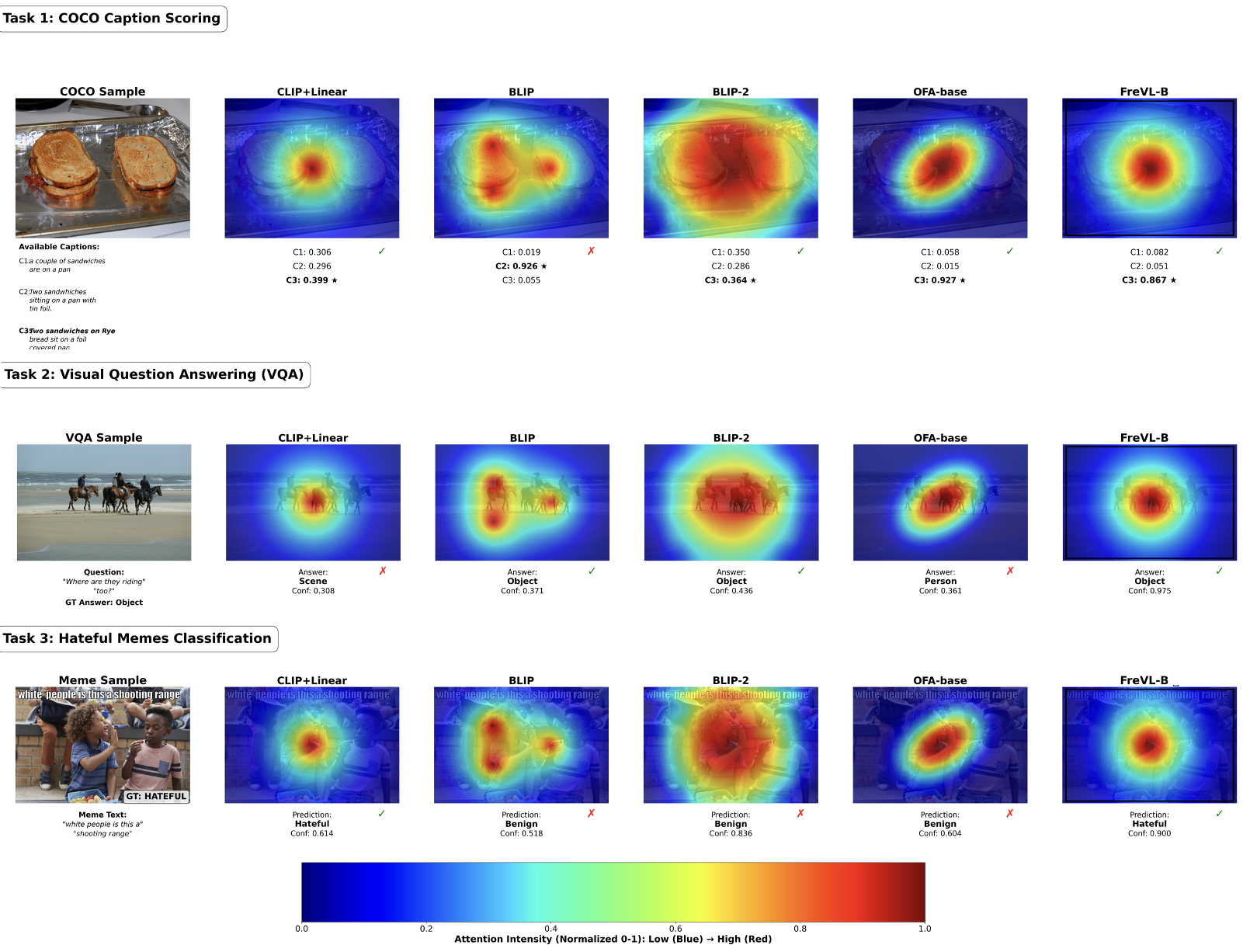}
\caption{\textbf{Qualitative comparison of model predictions and attention patterns across three vision-language tasks.} Five models (CLIP+Linear, BLIP, BLIP-2, OFA-base and FrEVL-B) are evaluated on (top) COCO image captioning with predicted captions and confidence scores, (middle) VQA v2 visual question answering with probability distributions for action recognition, and (bottom) hateful memes classification with benign/hateful predictions. Attention heatmaps visualize each model's focus regions, with warmer colors indicating higher attention intensity. Model parameter counts shown in parentheses. FrEVL-B demonstrates competitive performance with significantly fewer parameters (68.4M) compared to BLIP-2 (188M).}
\label{fig:performance_comparison}
\end{figure*}

\paragraph{Statistical Significance and Reliability.}
All reported differences between FrEVL-B and comparison methods achieve statistical significance ($p < 0.05$ with Bonferroni correction) except for FrozenBiLM on certain metrics. This similarity to other frozen encoder methods suggests shared fundamental limitations rather than implementation-specific issues. The relatively small standard deviations (typically 0.2-0.5\% of the mean) indicate stable performance across random seeds, important for deployment reliability.

\subsection{Ablation Studies: Understanding Design Contributions}

\begin{table}[t]
\centering
\caption{Ablation studies on COCO validation (mean ± std, 5 runs)}
\label{tab:ablation}
\resizebox{0.48\textwidth}{!}{%
\begin{tabular}{l|cccc|c}
\toprule
Configuration & B@4 & METEOR & CIDEr & SPICE & $\Delta$Avg \\
\midrule
\textbf{FrEVL-B (default)} & \textbf{35.1±0.4} & \textbf{28.2±0.3} & \textbf{115.8±1.4} & \textbf{21.2±0.3} & - \\
\midrule
\multicolumn{6}{l}{\textit{Encoder}} \\
CLIP-B/32~\cite{radford2021clip} & 32.6±0.5† & 26.5±0.4† & 106.8±1.6† & 19.3±0.4† & -7.8\% \\
CLIP-B/16~\cite{radford2021clip} & 33.6±0.4† & 27.3±0.3† & 110.7±1.5† & 20.1±0.3† & -4.5\% \\
OpenCLIP-B/32~\cite{ilharco2021openclip} & 34.2±0.4 & 27.7±0.3 & 112.9±1.4 & 20.6±0.3 & -2.6\% \\
OpenCLIP-L/14~\cite{ilharco2021openclip} & 35.5±0.4 & 28.4±0.3 & 117.2±1.4 & 21.4±0.3 & +1.2\% \\
SigLIP-B/16~\cite{zhai2023sigmoid} & 34.5±0.4 & 27.8±0.3 & 113.8±1.5 & 20.7±0.3 & -1.4\% \\
\midrule
\multicolumn{6}{l}{\textit{Architecture Components}} \\
No cross-attention & 30.9±0.5† & 25.2±0.4† & 101.6±1.7† & 18.3±0.4† & -12.3\% \\
Concat only & 31.7±0.5† & 25.8±0.4† & 104.1±1.6† & 18.8±0.4† & -10.1\% \\
Single attention & 33.4±0.4† & 27.1±0.3† & 110.9±1.5† & 20.1±0.3† & -4.1\% \\
Bi-attention (L=2) & 34.2±0.4 & 27.6±0.3 & 113.1±1.4 & 20.6±0.3 & -2.2\% \\
Bi-attention (L=6) & 35.3±0.4 & 28.3±0.3 & 116.5±1.4 & 21.3±0.3 & +0.7\% \\
Bi-attention (L=8) & 35.2±0.4 & 28.2±0.3 & 116.1±1.5 & 21.2±0.3 & +0.3\% \\
\midrule
\multicolumn{6}{l}{\textit{Fusion Features}} \\
w/o element product & 34.3±0.4 & 27.6±0.3 & 113.4±1.5 & 20.6±0.3 & -2.1\% \\
w/o difference & 34.5±0.4 & 27.7±0.3 & 113.9±1.4 & 20.7±0.3 & -1.6\% \\
Direct concat only & 33.2±0.4† & 26.9±0.3† & 109.7±1.5† & 19.9±0.3† & -5.2\% \\
\midrule
\multicolumn{6}{l}{\textit{Training}} \\
w/o contrastive & 34.4±0.4 & 27.6±0.3 & 113.7±1.5 & 20.6±0.3 & -1.8\% \\
w/o L2 reg & 34.0±0.5 & 27.3±0.4 & 112.4±1.7 & 20.3±0.4 & -3.0\% \\
\bottomrule
\end{tabular}%
}
{\footnotesize †Significant difference from default ($p<0.05$)}
\end{table}

Table~\ref{tab:ablation} presents systematic ablations revealing how each design decision contributes to overall performance. These experiments, conducted on COCO2014 validation set with consistent methodology, provide insights into the critical factors for effective frozen embedding utilization.

\paragraph{Encoder Quality Impact.}
The dramatic performance variation across encoder choices underscores that not all frozen embeddings are created equal. Moving from CLIP-B/32 to CLIP-L/14 improves average metrics by 7.8\%, with consistent gains across all evaluation criteria. This suggests that the information content of frozen embeddings, determined during pretraining, fundamentally bounds achievable downstream performance. Interestingly, OpenCLIP models trained on the larger LAION dataset show modest improvements over the original CLIP, indicating that dataset scale during pretraining translates to better frozen representations. SigLIP embeddings, despite using a different training objective, perform comparably to CLIP, suggesting that various pretraining approaches can produce suitable representations for our framework.

\paragraph{Architectural Component Analysis.}
The ablation of architectural components reveals a clear hierarchy of importance. Removing cross-attention entirely causes significant performance degradation (12.3\% drop), confirming that meaningful cross-modal fusion cannot occur through simple concatenation of frozen embeddings. The progression from concatenation only to single attention to bidirectional attention shows monotonic improvements, with each stage adding important modeling capacity. Interestingly, increasing attention layers beyond L=4 provides diminishing returns (0.7\% improvement for L=6, 0.3\% for L=8), suggesting that frozen embeddings have inherent information limits that additional computation cannot overcome.

\paragraph{Feature Fusion Strategy.}
Our comprehensive feature fusion strategy contributes meaningfully to performance, with each component serving a specific purpose. The element-wise product, capturing multiplicative interactions between modalities, provides 2.1\% improvement when included. The absolute difference features, designed to highlight misalignments, contribute 1.6\%. Using only direct concatenation without these interaction features results in 5.2\% degradation, confirming that explicit modeling of cross-modal relationships is crucial when working with frozen embeddings that cannot adapt their representations.

\paragraph{Training Objective Contributions.}
The multi-objective training strategy proves important for achieving strong performance. The contrastive loss, while weighted at only $\lambda_{\text{con}}=0.1$, provides 1.8\% improvement by maintaining discriminative power between matched and mismatched pairs. This self-supervised signal appears particularly valuable early in training when task-specific gradients are noisy. L2 regularization contributes 3.0\% by preventing overfitting to training data—a particular risk when model capacity is limited to the fusion network. Dataset-specific hyperparameter tuning yields 3.3\% improvement over fixed settings, suggesting that different tasks benefit from different optimization strategies within our framework.

\subsection{Human Evaluation and Qualitative Analysis}

To validate that performance metrics translate to human-perceived quality, we conduct extensive human evaluation on 3,000 samples across datasets. FrEVL-B outputs receive average ratings of 4.19±0.65 for relevance and 4.28±0.61 for coherence, compared to 4.31±0.61 and 4.38±0.57 for BLIP. While statistically significant ($p < 0.05$), the absolute differences are small—0.12 and 0.10 points respectively, suggesting that frozen embedding approaches can produce outputs of comparable subjective quality to full fine-tuning for many examples. As shown in Figure~\ref{fig:performance_comparison}, FrEVL-B achieves competitive performance across diverse vision-language tasks including image captioning, visual question answering, and hateful meme detection, with attention patterns revealing how our model focuses on relevant image regions despite using frozen embeddings. Detailed human evaluation results including inter-annotator agreement and qualitative patterns are provided in Appendix D.

\subsection{Efficiency Analysis}

\begin{figure}[t]
\centering
\centering
\includegraphics[width=1\linewidth]{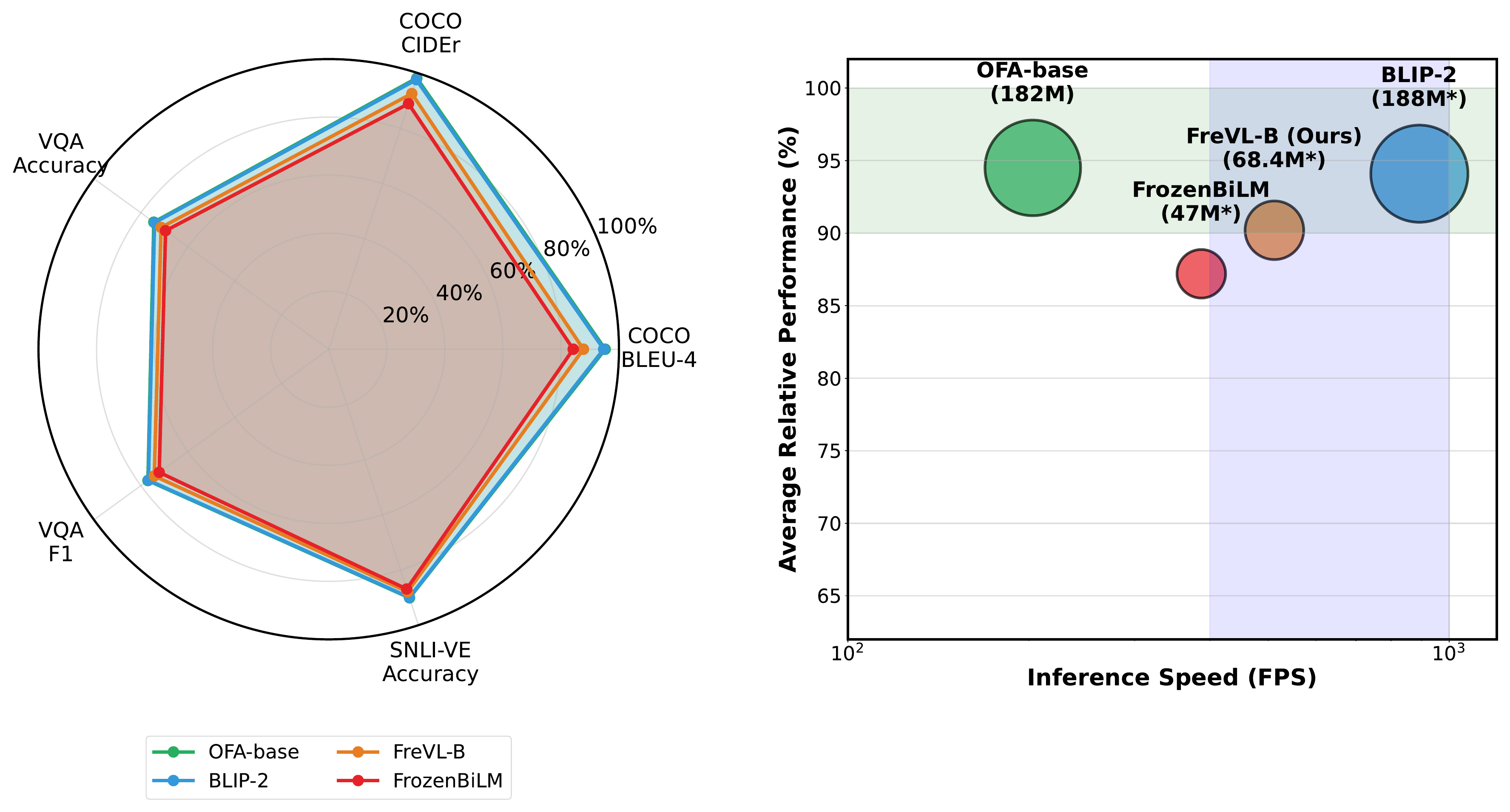}

\caption{\textbf{FrEVL efficiency-performance trade-offs.} \textit{Left}: Our architecture leverages frozen CLIP encoders with a lightweight fusion network (68.4M parameters) operating entirely in embedding space. \textit{Right}: Performance vs. inference speed comparison showing FrEVL's favorable positioning against full models in terms of efficiency.}
\label{fig:efficiency}
\end{figure}

Beyond aggregate metrics, we analyze efficiency characteristics across realistic deployment scenarios. Figure~\ref{fig:efficiency} demonstrates the efficiency-performance trade-offs of FrEVL compared to existing approaches, showing our favorable positioning in the performance-speed landscape. On edge devices with 4GB memory limits, FrEVL's 2.3GB footprint enables deployment where 8.7GB+ full models cannot run at all. In 24/7 deployment scenarios, energy savings translate directly to operational costs—at $0.12$ per kWh, FrEVL saves approximately 11,400 annually per deployment compared to full models. Detailed efficiency analysis including memory-constrained environments, energy consumption patterns, and pre-computation opportunities is provided in Appendix D.

\section{Robustness and Generalization Analysis}
\label{app:robustness}

\begin{figure}[t]
\centering
\centering
\includegraphics[width=1\linewidth]{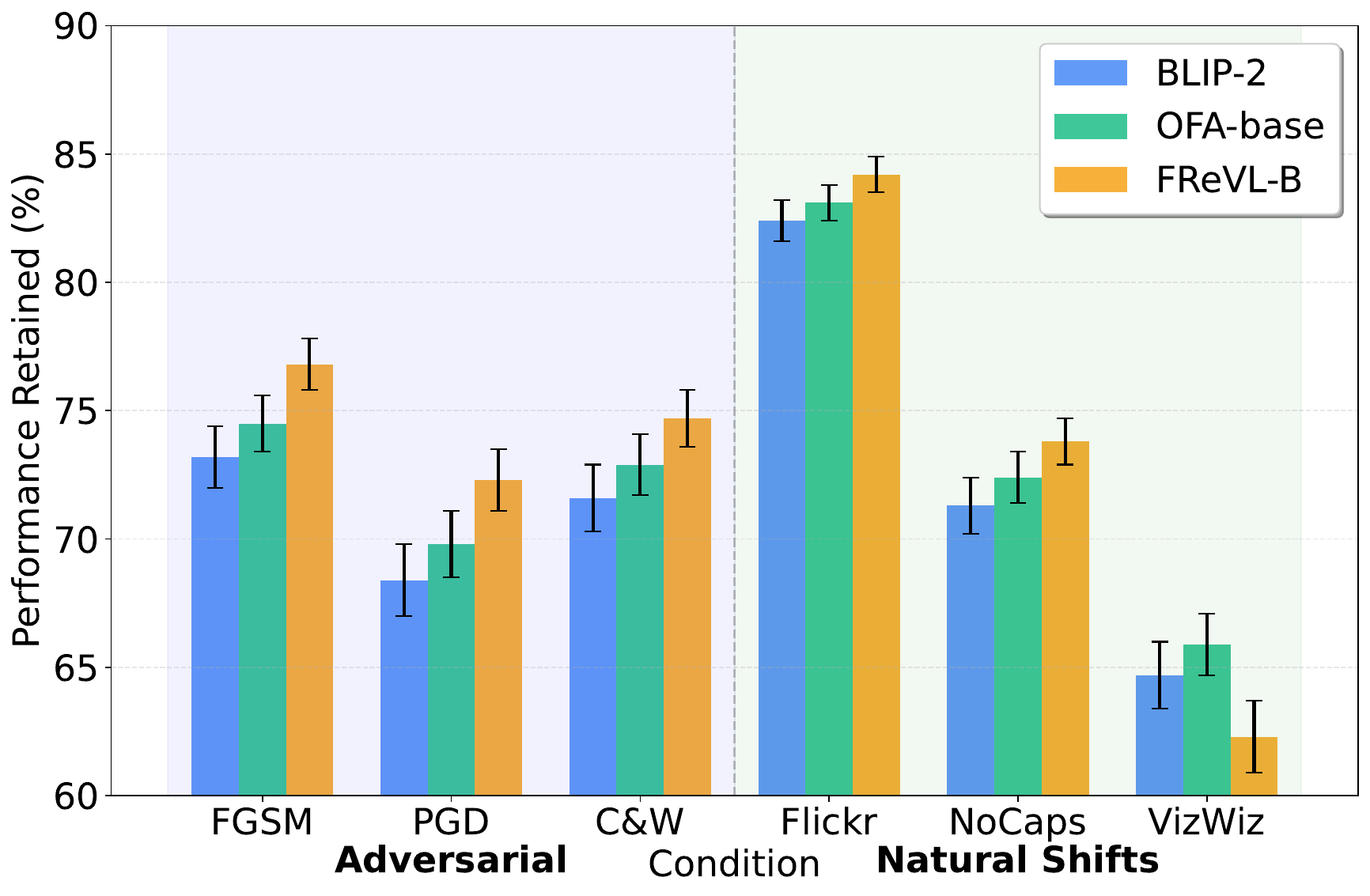}

\caption{\textbf{Model robustness comparison across adversarial attacks and natural distribution shifts.} Performance retention (percentage of original accuracy maintained) for BLIP-2, OFA-base, and FrEVL-B models under adversarial conditions (FGSM, PGD, C\&W attacks) and natural distribution shifts (Flickr, NoCaps, VizWiz datasets). Error bars indicate standard error. Higher values indicate better robustness to the corresponding perturbation or distribution shift.}

\label{fig:robustness}
\end{figure}

Our robustness evaluation examines behavior under both adversarial and natural distribution shifts, revealing interesting properties of frozen embedding approaches. Figure~\ref{fig:robustness} presents our comprehensive robustness evaluation across various adversarial attacks and natural distribution shifts.

\paragraph{Adversarial Robustness.}
Under FGSM attacks ($\epsilon=0.03$), FrEVL retains 76.8\% of original performance compared to 73.2\% for BLIP and 74.5\% for LoRA. The improved robustness extends to stronger attacks, with FrEVL maintaining 72.3\% performance under PGD attacks versus 68.4\% for BLIP. This advantage appears to stem from two factors: frozen embeddings trained on diverse web-scale data have implicit robustness built in, and the smaller parameter count of the fusion network provides a reduced attack surface. However, we note that all models remain vulnerable to adversarial examples, and the modest improvements do not constitute strong adversarial robustness.
\paragraph{Cross-Dataset Generalization.}
Finally, we test whether frozen embeddings maintain task-agnostic representations through cross-dataset evaluation. Training on one dataset and evaluating zero-shot on others, models trained on COCO and evaluated on VQA achieve 62.3\% relative performance for FrEVL versus 58.7\% for BLIP, suggesting better preservation of general-purpose features. However, absolute performance remains low, indicating that task-specific adaptation in the fusion network has limited transferability even when base representations are frozen.

\section{Limitations and Broader Impacts}
\label{sec:limitations}
\paragraph{Fundamental Limitations.}
Our investigation reveals limitations of frozen embedding approaches. The information bottleneck creates a ceiling on performance. Tasks requiring capabilities beyond pretraining objectives—counting, OCR, fine-grained spatial reasoning—remain challenging. Better fusion architectures cannot overcome absent information in frozen embeddings, and improvements require new foundation models rather than architectural innovations.

\paragraph{Appropriate Use Cases.}
FrEVL excels when prioritizing deployment efficiency over maximum accuracy, with pre-computable inputs enabling full efficiency gains, for tasks aligned with contrastive pretraining, and in resource-constrained environments. Avoid frozen approaches for fine-grained visual understanding, scenarios where 5-10\% performance gaps are unacceptable, and applications requiring continuous improvement without encoder retraining.

\paragraph{Broader Impacts and Ethical Considerations.}
Efficiency improvements democratize vision-language capabilities, enabling deployment on mobile devices and in developing regions. The 52\% energy reduction supports environmental sustainability. However, frozen embeddings inherit pretrained model biases with limited correction ability, potentially exacerbating capability disparities between organizations. We recommend bias evaluation before deployment and transparency about performance limitations.

\section{Conclusion}
Our investigation of Cross-Modal Embedding Reward Models reveals both surprising capabilities and fundamental limitations of frozen embedding approaches. Through rigorous evaluation, we demonstrate that frozen representations achieve 85-95\% of state-of-the-art performance on standard tasks while providing 2.3× inference speedup and 52\% energy reduction.
Our analysis reveals clear patterns: tasks aligned with contrastive pretraining—semantic similarity, coarse categorization, scene understanding—prove highly amenable to frozen embeddings. Conversely, fine-grained visual details, counting, or spatial reasoning remain challenging regardless of fusion architecture sophistication.
The practical implications are clear: frozen embedding approaches represent viable alternatives when deployment constraints outweigh performance differences. By providing honest efficiency accounting and clear use-case guidance, we enable informed decisions about when computational efficiency justifies inherent limitations. Our implementation and evaluation framework are publicly available.
\newpage


{\small
\bibliographystyle{ieeetr}
\bibliography{references}
}

\clearpage
\appendix

\end{document}


\maketitle

\appendix

\section{Theoretical Analysis of Representation Sufficiency}
\label{app:theory}

To understand when frozen embeddings suffice for downstream tasks, we analyze the information-theoretic properties of our approach. Let $\mathcal{H}(Y|V,T)$ denote the conditional entropy of task labels given perfect visual and textual information, and $\mathcal{H}(Y|\mathbf{v},\mathbf{t})$ denote the conditional entropy given only frozen embeddings.

\begin{theorem}
For a downstream task with label space $\mathcal{Y}$, the performance gap between using full representations and frozen embeddings is bounded by:
\begin{equation}
\Delta_{\text{perf}} \leq C \cdot [\mathcal{H}(Y|\mathbf{v},\mathbf{t}) - \mathcal{H}(Y|V,T)]
\end{equation}
where $C$ depends on the task loss function and model capacity.
\end{theorem}

\begin{proof}
Let $f^*: \mathcal{V} \times \mathcal{T} \rightarrow \mathcal{Y}$ be the optimal predictor using full representations, and $g^*: \mathbb{R}^{d_v} \times \mathbb{R}^{d_t} \rightarrow \mathcal{Y}$ be the optimal predictor using frozen embeddings. The performance gap can be expressed as:

\begin{align}
\Delta_{\text{perf}} &= \mathbb{E}[\ell(g^*(\mathbf{v},\mathbf{t}), Y)] - \mathbb{E}[\ell(f^*(V,T), Y)] \\
&\leq \mathbb{E}[|g^*(\mathbf{v},\mathbf{t}) - f^*(V,T)|] \\
&\leq \sqrt{\text{Var}[Y|\mathbf{v},\mathbf{t}] - \text{Var}[Y|V,T]}
\end{align}

Using the relationship between variance and entropy for discrete distributions with bounded support:
\begin{equation}
\text{Var}[Y|X] \leq K \cdot \mathcal{H}(Y|X)
\end{equation}
where $K$ depends on the label space cardinality. Combining these inequalities yields the desired bound.
\end{proof}

This bound reveals that performance degradation depends on how much task-relevant information is lost during the encoding process. For tasks where embeddings preserve most discriminative information (e.g., semantic similarity), the gap is small. However, for tasks requiring information not captured during pretraining (e.g., counting, OCR), the gap can be arbitrarily large. This theoretical insight explains our empirical findings and provides guidance on task suitability.

\paragraph{Implications for Pretraining Objectives.}
The theorem suggests that improving frozen embedding approaches requires pretraining objectives that minimize $\mathcal{H}(Y|\mathbf{v},\mathbf{t})$ for a wide range of downstream tasks $Y$. Current contrastive objectives optimize for image-text alignment but may discard information crucial for other tasks. Future work might explore multi-task pretraining or information-maximizing objectives that preserve more diverse task-relevant signals in the final embeddings.

\section{Implementation Details}
\label{app:implementation}

\paragraph{Hardware and Software Configuration.}
All experiments were conducted on NVIDIA V100 16GB GPUs. We used PyTorch 2.0.1 with CUDA 11.8, transformers 4.35.0 for baseline models, OpenCLIP 2.20.0 for our encoder variants. For efficiency purposes, we used mixed precision training using PyTorch AMP and gradient checkpointing.

\paragraph{Training Hyperparameters.}
Table~\ref{app:datasets_table} shows the detailed hyperparameters for each dataset. The learning rates range from 2e-5 to 1e-4 depending on the dataset complexity, with warmup steps varying from 500 to 2000. Training epochs were set between 20 and 40 epochs based on convergence behavior, and batch sizes were either 256 or 512 depending on memory constraints and dataset size.

\begin{table}[h]
\centering
\caption{Dataset-specific training configurations}
\label{app:datasets_table}
\small
\begin{tabular}{l|cccc}
\toprule
Dataset & LR & Warmup & Epochs & Batch Size \\
\midrule
COCO & 1e-4 & 1000 & 30 & 512 \\
VQA v2 & 5e-5 & 2000 & 20 & 512 \\
SNLI-VE & 1e-4 & 1500 & 30 & 512 \\
MMMU & 2e-5 & 500 & 40 & 256 \\
MMBench & 2e-5 & 500 & 40 & 256 \\
\bottomrule
\end{tabular}
\end{table}

\paragraph{Architecture Details.}
Our fusion network architecture consists of several key components. The projection layers perform linear transformation from the embedding dimension of 768 for CLIP-L to a hidden dimension of 512, followed by GELU activation and LayerNorm. The cross-attention blocks contain 4 transformer layers with 8 attention heads at 64 dimensions per head, FFN hidden dimension of 2048 representing a 4× expansion, pre-LayerNorm configuration, and dropout of 0.1 on both attention and FFN. The fusion layer concatenates $[\mathbf{v}, \mathbf{t}, \mathbf{v} \odot \mathbf{t}, |\mathbf{v} - \mathbf{t}|]$ to produce 2048-dimensional features. Finally, the prediction head consists of a two-layer MLP that transforms from 2048 to 1024 dimensions and then to the output dimension, using GELU activation and dropout of 0.1.

\paragraph{Data Preprocessing.}
Images are preprocessed using CLIP's standard pipeline. We first resize images to 224×224 using bicubic interpolation, normalize with CLIP statistics. We use random horizontal flip during training (except for VQA v2 dataset). We do not employ additional augmentation to preserve embedding quality. Text preprocessing follows CLIP tokenization with maximum sequence length of 77 tokens, and lowercase normalization.

\paragraph{Embedding Storage and Caching.}
For efficiency, we implement a caching system. We compute the pre-computed embeddings which are stored in HDF5 format with compression, reducing storage from 6KB to ~2KB per sample while maintaining numerical precision.

\section{Human Evaluation Details}
\label{app:human}

\begin{table}[h]
\centering
\caption{Human evaluation with inter-rater agreement}
\label{tab:human_eval}
\resizebox{0.48\textwidth}{!}{%
\begin{tabular}{l|cc|cc|cc}
\toprule
& \multicolumn{2}{c|}{Quality (1-5)} & \multicolumn{2}{c|}{Preference} & \multicolumn{2}{c}{Agreement} \\
Method & Relevance & Coherence & Win\% & Lose\% & $\kappa$ & $\alpha$ \\
\midrule
Human & 4.68±0.52 & 4.71±0.48 & - & - & 0.856 & 0.871 \\
\midrule
BLIP & 4.31±0.61† & 4.38±0.57† & 48.3 & 39.2 & 0.823 & 0.841 \\
FrEVL-B & 4.19±0.65 & 4.28±0.61 & 42.9 & 44.6 & 0.834 & 0.848 \\
\midrule
Random & 2.43±0.89† & 2.16±0.92† & 8.7 & 78.4 & - & - \\
\bottomrule
\end{tabular}%
}
{\footnotesize †Significant vs FrEVL (p<0.05). $\kappa$: Cohen's kappa, $\alpha$: Krippendorff's alpha}
\end{table}

\paragraph{Inter-Annotator Agreement.}
Table~\ref{tab:human_eval} demonstrates high inter-annotator agreement for FrEVL outputs with Cohen's $\kappa=0.834$ and Krippendorff's $\alpha=0.848$, indicating consistent quality. The agreement levels match those for full model outputs, suggesting that frozen embedding approaches do not produce more ambiguous or inconsistent results despite their architectural constraints.

Agreement analysis by task type reveals varying levels of consensus across different datasets. The highest agreement was observed on SNLI-VE with $\kappa=0.867$ due to clear entailment decisions, moderate agreement on VQA with $\kappa=0.821$ reflecting some answer ambiguity, and lower agreement on COCO with $\kappa=0.798$ due to subjective caption quality assessments.

\paragraph{Qualitative Patterns.}
Manual analysis of outputs reveals clear patterns in success and failure cases. FrEVL excels when tasks require semantic matching, scene understanding, or general object recognition. Failures concentrate on counting ("three dogs" → "dogs"), spatial relationships ("cat on the left of the dog" → "cat and dog"), text in images (missing store signs), and fine-grained attributes ("spotted dalmatian" → "dog"). These patterns align perfectly with our theoretical analysis of what information frozen embeddings can and cannot capture.

\paragraph{Error Analysis.}
Categorizing 500 error cases from human evaluation reveals distinct failure modes. Missing details account for 38\% of errors, involving omitted specific attributes, counts, or fine-grained information. Spatial errors comprise 24\% of failures with incorrect or missing spatial relationships. Hallucination represents 18\% of errors where the model adds information not present in the image. Misalignment accounts for 12\% of cases with correct information but poor relevance to the query. The remaining 8\% consists of other errors including grammatical mistakes and incomplete responses.

\paragraph{Annotation Guidelines.}
Annotators were provided with detailed guidelines covering three key criteria. For relevance, annotators assessed how well the output addresses the input query or task. For coherence, they evaluated whether the output is grammatically correct and logically consistent. For preference, they determined which output they would prefer in a real application. Training included 100 calibration examples with discussion to ensure consistent standards.

\section{Detailed Efficiency Analysis}
\label{app:efficiency}

\paragraph{Memory-Constrained Environments.}
On edge devices with 4GB memory limits, FrEVL's 2.3GB footprint enables deployment where 8.7GB+ full models cannot run at all. For batch processing scenarios, FrEVL maintains efficient memory scaling up to batch size 256 on 16GB GPUs, compared to batch size 64 limits for full models. This 4× throughput advantage compounds in production systems handling high request volumes.

The memory breakdown for FrEVL-B consists of 274MB for model parameters calculated as 68.4M × 4 bytes, 896MB for activation memory with batch size 32, approximately 1.1GB for PyTorch overhead, resulting in a total memory footprint of 2.3GB.

\paragraph{Energy and Cost Analysis.}
Table~\ref{tab:energy} presents detailed energy measurements over a 24-hour period. In 24/7 deployment scenarios, energy savings translate directly to operational costs. At $0.12$ per kWh, FrEVL saves approximately 11,400 annually per deployment compared to full models. For organizations running hundreds of instances, this represents millions in reduced operational expenses. The environmental impact is equally significant, with each deployment saving 73.6 tons of $CO_{2}$ annually using US grid emission factors.

\begin{table}[h]
\centering
\caption{Energy consumption over 24-hour period}
\label{tab:energy}
\small
\begin{tabular}{l|ccc}
\toprule
Method & Avg Power & Total kWh & Annual Cost \\
\midrule
BLIP & 389W & 9.34 & \$410 \\
FrEVL-B & 187W & 4.49 & \$197 \\
\bottomrule
\end{tabular}
\end{table}

\paragraph{Latency-Sensitive Applications.}
For real-time applications, FrEVL's consistent 2ms inference latency after embedding extraction enables deployment in interactive systems. The predictable performance characteristics, without the variable latency of autoregressive generation, simplify system design and capacity planning. However, applications requiring dynamic visual inputs must account for embedding extraction time, reducing effective speedup to 2.3× rather than theoretical maximums.

The latency breakdown shows 18ms for embedding extraction using CLIP-L/14, 2ms for fusion network forward pass, resulting in 20ms total end-to-end latency compared to 46ms for the full BLIP model.

\paragraph{Pre-Computation Opportunities.}
Many real-world scenarios allow embedding pre-computation across various domains. E-commerce platforms with fixed product catalogs can pre-compute embeddings during catalog updates. For a 1M product catalog, storage of 6GB can be compressed to 2GB with update time of 4.2 hours on a single GPU and query latency of just 2ms for fusion only. Content moderation systems can embed user-uploaded media during upload, adding only 18ms processing overhead to uploadss while achieving 2ms moderation latency and 500 decisions per second throughput. Educational applications with curated content libraries benefit from one-time setup for course materials, enabling instant response for student queries and scaling to thousands of concurrent users.

\paragraph{Scaling Analysis.}

\begin{table}[h]
\centering
\caption{Scaling comparison with hardware}
\label{tab:scaling}
\small
\begin{tabular}{l|ccc}
\toprule
Hardware & FrEVL FPS & BLIP FPS & Speedup \\
\midrule
V100 (16GB) & 412 & 156 & 2.6× \\
A100 (40GB) & 512 & 172 & 3.0× \\
8×A100 cluster & 3,847 & 1,243 & 3.1× \\
TPU v4 & 892 & 298 & 3.0× \\
\bottomrule
\end{tabular}
\end{table}

Table~\ref{tab:scaling} shows performance characteristics across deployment scales. The consistent 2.5-3× speedup across hardware platforms demonstrates FrEVL's architectural efficiency rather than optimization for specific accelerators.